\documentclass[]{ceurart}

\usepackage{fvextra}
\usepackage[scale=0.85]{sourcecodepro}
\input{pyg-tango.tex}
\DefineVerbatimEnvironment{MintedVerbatim}{Verbatim}%
  {breaklines=true,fontsize=\footnotesize,vspace=-10.2pt}
\usepackage{float}
\newfloat{listing}{tbp}{lol}
\newcommand{\listingscaption}{Listing}
\floatname{listing}{\listingscaption}
\usepackage{newfloat}
\SetupFloatingEnvironment{listing}{name=Listing}
\floatstyle{plaintop}
\restylefloat{listing}
\usepackage{caption}
\usepackage{microtype}

\begin{document}

\copyrightyear{2026}
\copyrightclause{Copyright for this paper by its authors. Use permitted under Creative Commons License Attribution 4.0 International (CC BY 4.0).}

\conference{Third Workshop on Knowledge Graphs and Neurosymbolic AI (KG-NeSy 2026), in ISWC 2026 Workshops Joint Proceedings, October 25--26, 2026, Bari, Italy}

\title{Ontology-Mediated Neurosymbolic Constraint Acquisition from Multiple Stakeholders}

\author{Stefan Bischof}[%
orcid=0000-0001-9521-8907,
email=bischof.stefan@siemens.com,
]
\author{Juliana Kainz}[%
orcid=0000-0002-1293-0315,
email=juliana.kainz@siemens.com,
]
\author{Danilo Valerio}[%
orcid=0000-0003-4104-3974,
email=danilo.valerio@siemens.com,
]
\address{Siemens AG Österreich, Vienna, Austria}

\begin{abstract}
Neurosymbolic research typically assumes a pre-existing symbolic specification, leaving the upstream challenge of acquiring and formalizing requirements and constraints largely unaddressed. We present an architecture that fills this gap by using an OWL configuration ontology to mediate between neural constraint sources and downstream consumers. In this framework, LLM assistants elicit soft stakeholder preferences, while hardware specifications define hard physical and engineering limits. The ontology unifies these heterogeneous inputs, leverages description logic to identify unsatisfiability, and generates symbolic explanations that enable LLMs to interactively renegotiate terms with users. Any remaining conflicts are resolved downstream via priority-based relaxation. We illustrate our approach on a microgrid use case from the FLEXI project and argue its generalizability to multi-stakeholder domains where constraint acquisition is distributed across human and automated sources of unequal authority.
\end{abstract}


\maketitle

\section{Introduction}

Neurosymbolic systems can be categorized by how they couple neural components with symbolic specifications \cite{boxology}. Most existing literature focuses heavily on the downstream problem: ensuring that neural outputs comply with a given specification.
Typically a neural component proposes a candidate output (an answer, a structured string, a proof) and a
symbolic layer checks it against logical constraints, rejecting or repairing
violations. This paradigm has driven significant advances in verification,
constrained decoding, and symbolic verifiers that check neural
outputs~\cite{picard,logiclm,linc,satlm}.

A much less explored challenge is the upstream problem of how these symbolic specifications are actually constructed. In many real settings the constraint set is neither fixed nor given by a
single authority. It is acquired, assembled from heterogeneous sources whose
constraints may be mutually inconsistent. Physical and safety limits come from
hardware specifications and are non-negotiable. Other constraints are preferences:
what a particular user wants, expressed in natural language and elicited through
dialogue. Preferences originate from multiple stakeholder groups whose interests may
conflict and whose claims do not carry equal weight; for example, when operating an energy grid, an operator's grid-stability
requirement outranks an individual user's preference.
Before the specification can be used, these sources must be reconciled into a single,
consistent constraint set, and the conflicts must be detected and resolved. %

In this position paper, we argue that a configuration ontology is an ideal
mediating layer for this upstream acquisition challenge, and outline the
architecture it enables. Our work builds on CONTO, a
configuration ontology based on the Web Ontology Language (OWL), developed at Siemens
for vendor-independent
product configuration~\cite{conto-semantics25}. CONTO already provides a
standards-based representation of components, features, and constraints, together
with Description-Logic consistency checking. However it is limited to modeling \emph{hard
constraints only}: every constraint must be satisfied, and there is no notion of
preference, priority, or optimization objective. Overcoming this limitation is what
multi-stakeholder constraint acquisition demands, and addressing it forms the core of our
approach.
Concretely, we make the following contributions:
\begin{itemize}
  \item We frame \textbf{multi-stakeholder constraint acquisition} as a distinct
    neurosymbolic problem, the necessary complement to the
    well-studied downstream validation task.
  \item We propose an \textbf{ontology-mediated architecture} that integrates LLM assistants and hardware sources into an OWL configuration ontology. By extending this ontology with a hard/soft distinction and priority levels, user preferences become relaxable, weighted constraints alongside hard physical limits (Section~\ref{sec:approach}).
  \item We describe a \textbf{closed neurosymbolic loop} in which description-logic
    reasoning detects structural conflicts and returns symbolic explanations that
    drive the assistants' renegotiation with users, and we instantiate the
    architecture on the microgrid (Section~\ref{sec:usecase}).
\end{itemize}

Throughout, we ground the architecture in the microgrid use case of the FLEXI
project,\footnote{\url{https://flexi-cetp.eu/}}
a business park with
photovoltaics, stationary batteries, and electric-vehicle (EV) chargers whose charging and storage
must be scheduled for sustainability, grid stability, and cost. Its constraints
come from sources of unequal authority: EV owners state charging preferences, a
charge-point operator sets infrastructure requirements, and hardware
specifications fix the hard physical limits.
Two requests show what makes reconciliation hard: charging 90\,kWh into a battery
that holds 60 can never be satisfied at all, whereas a full charge by 16:00 is
feasible on its own yet may be unschedulable once many vehicles charge at once.

\section{Related Work}
\label{sec:related}

We position our approach against three strands of work.

The dominant neurosymbolic pattern in constrained settings uses symbolic knowledge
to constrain or verify neural outputs: constrained decoding restricts generation to
a formal grammar~\cite{picard}, and symbolic solvers or provers check a proposed
solution and feed errors back to the model~\cite{linc,satlm}. These approaches assume the
constraint specification is given; our work targets the prior question of how a
consistent specification is acquired in the first place. Logic-LM~\cite{logiclm} is
the closest structural analogue, iterating generation, symbolic checking, and
revision, but its feedback corrects the \emph{formalization} of a single given
problem, whereas ours reconciles \emph{competing constraints} elicited from
multiple stakeholders.

There is a substantial constraint-programming literature on \emph{constraint
acquisition}, learning a constraint network from examples of valid and invalid
assignments~\cite{constraint-acquisition}. This includes active methods that acquire
constraints through membership queries to a user~\cite{quacq}. This paradigm typically
assumes a single implicit authority from which constraints are learned via structured data or query responses. It
rarely assumes elicitation through natural-language dialogue, leverages an explicit ontology
as the shared representation, or manages multiple stakeholder groups with differing
priority.

Recent work combines LLMs with ontologies and knowledge graphs for requirements
elicitation and configuration: conversational ontology engineering guides users
from informal input to structured requirements~\cite{ontochat}, and LLMs paired
with constraint programming support interactive preference
elicitation~\cite{lawless-elicitation}. Closely related, a configuration copilot
integrates an LLM with a constraint solver for interactive product
configuration~\cite{conto-copilot}. We build on this line but shift the emphasis
from authoring a single product model to reconciling constraints acquired
concurrently from several conflicting sources of unequal authority.

\section{Approach}
\label{sec:approach}

We present the architecture in two steps: first the mediating layer itself, the
sources that feed it, and the vocabulary extension that makes stakeholder
preferences relaxable; then the consistency check this enables and the feedback
loop it drives.

\subsection{Architecture Overview}

\begin{figure}[t]
    \centering
    \includegraphics[width=1\linewidth]{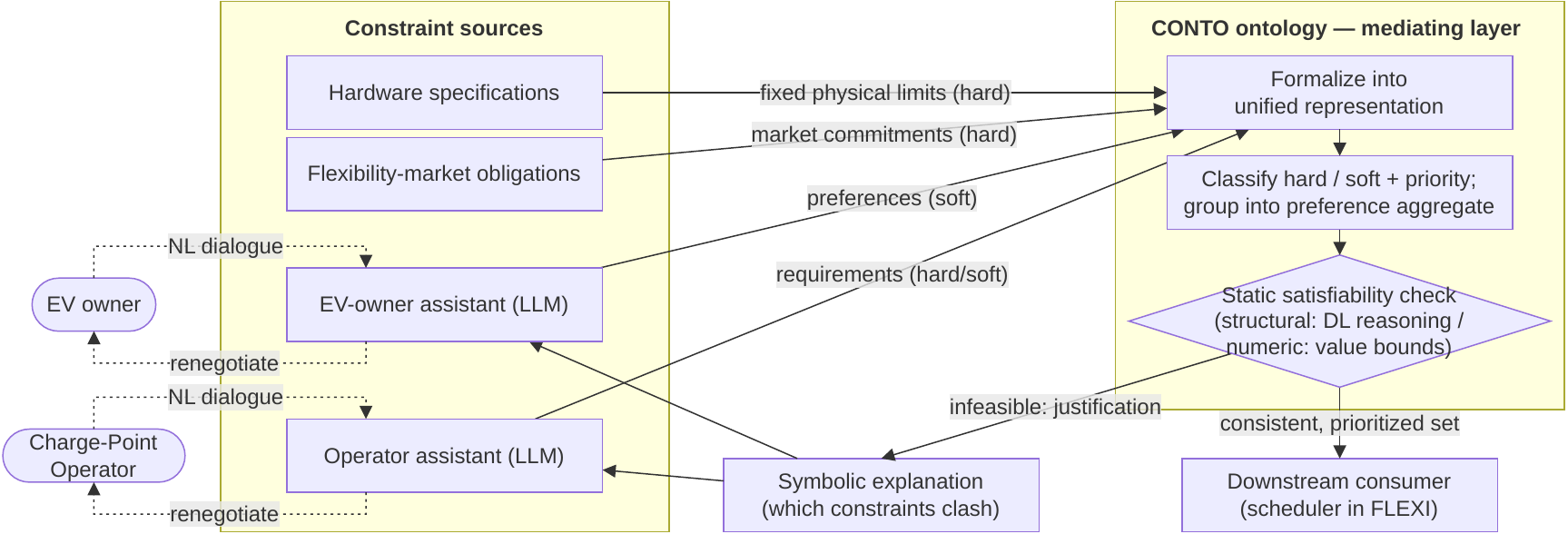}
  \caption{The constraint-acquisition loop, instantiated on the FLEXI microgrid.
    Constraint sources of unequal authority (left) feed the CONTO mediating layer
    (right), which this paper addresses; the downstream consumer (a scheduler in
    FLEXI) is out of scope.}
  \label{fig:architecture}
\end{figure}

We propose an architecture in which a configuration ontology sits between the
sources that supply constraints and the downstream consumer that uses them, shown
as the mediating layer of Figure~\ref{fig:architecture}.
Constraints enter from two kinds of source: LLM-based assistants that elicit
preferences from human stakeholders in dialogue, and hardware specifications that
contribute fixed physical limits. The ontology formalizes every incoming
constraint into a single representation, classifies it as hard or soft and
attaches a priority, and checks whether each soft preference can be satisfied at
all given the hard limits. A preference that cannot is turned into a
symbolic explanation and returned to the originating assistant, which renegotiates
with the user. Acquisition is continuous rather than a one-off batch: constraints may
arrive or change at any time, so each is checked as it arrives and, whenever the
feasible set changes, the scheduler is re-invoked. This paper concerns the acquisition
loop; the scheduler is downstream, and we treat it here as a consumer of the acquired
constraint set. It is in the acquisition loop that the neural and symbolic
components are coupled: the ontology mediates between them, mapping heterogeneous,
partially conflicting inputs to a single consistent and prioritized constraint set.

Our mediator builds on CONTO, a suite of OWL ontologies for vendor-independent
product configuration~\cite{conto-semantics25}. CONTO separates three layers: a
generic meta-model (L2) defining the vocabulary of components, features, and
constraints; a product line (L1) that defines a specific configurable artifact
against L2; and a configuration (L0), a concrete configured instance. It offers
two complementary formalizations of a product model. The \emph{Configuration
Vocabulary} (ConfigVoc, prefix \texttt{cv:}) is an instance-based OWL\,2\,QL
vocabulary in which a constraint is reified as data: a table constraint of
rows and cells, or a formula constraint carrying an expression tree. The
\emph{Configuration DL} ontology (ConfigDL, prefix \texttt{cdl:}) is a
class-based formalization in which a constraint \emph{is} an OWL axiom, so a
description-logic reasoner enforces it natively through consistency checking.

In both formalizations, however, every constraint is \emph{hard}: it must hold,
and there is no notion of a preference that may be relaxed, no priority among
constraints, and no optimization objective. For single-authority product
configuration this is adequate; for multi-stakeholder acquisition, where user
preferences must coexist with, and yield to, physical limits, it is the gap we
propose to close. Whether the preference semantics are better added to the
instance-based ConfigVoc (as additional data) or to the axiom-based ConfigDL
(where relaxable constraints do not map cleanly onto OWL's all-or-nothing axiom
semantics) is itself an open question we return to in
Section~\ref{sec:discussion}; for concreteness we illustrate the extension on
ConfigVoc below.

Each stakeholder group is served by its own instance of an LLM-based assistant,
configured for that group, which elicits
constraints in natural-language dialogue and maps them into the ontology's
vocabulary; because the assistants are the interface to human stakeholders, they
also carry back the explanations produced downstream (Section~\ref{sec:loop}). We
treat the elicited preferences as \emph{soft} constraints: statements of what a
stakeholder wants that should be satisfied where possible but may be relaxed. In
contrast, hardware and infrastructure specifications contribute \emph{hard}
constraints, fixed physical and safety limits that must always hold; these enter
the ontology directly, without elicitation, and are never relaxed.

\paragraph{Preferences as soft constraints.}
The central element of our approach is a small vocabulary extension that
introduces the hard/soft distinction, priorities, and preference aggregates.
We propose to classify each constraint as hard or soft, attach a numeric priority
(and, optionally, a weight) to soft constraints, and group a stakeholder's soft
constraints into a \emph{preference aggregate} that the downstream consumer should
try to satisfy. A hard constraint must hold: to \emph{violate} it renders the set
\emph{infeasible}. A soft constraint need not; leaving it unsatisfied is not a
violation but a \emph{relaxation} that incurs a \emph{violation cost} in proportion
to its priority, in the sense of valued and semiring-based
constraint-satisfaction problems (CSPs) and of weak constraints in answer set
programming (ASP)~\cite{vcsp,semiring-csp,weak-constraints}. Modeling preferences as first-class, weighted,
relaxable constraints, rather than as an out-of-band scoring function, keeps them
in the same formal representation as the hard limits, so that the same
satisfiability check and conflict explanation apply to both.

Listing~\ref{lst:constraints} illustrates the intended modeling in the FLEXI
setting, reusing existing ConfigVoc terms (prefix \texttt{cv:}) and introducing
the proposed preference vocabulary (prefix \texttt{pref:}): a hard physical limit
(charger power must not exceed fuse capacity), a prioritized soft preference
(the vehicle should reach its desired state of charge), and a preference aggregate
grouping that stakeholder's soft constraints. The formula bodies are
abbreviated for readability; ConfigVoc represents them as expression trees.

\begin{listing}
\caption{Modeling in the FLEXI microgrid: an existing ConfigVoc hard constraint
  (\texttt{cv:}) and the proposed preference extension (\texttt{pref:}) marking a
  soft, prioritized constraint and a preference aggregate. Formula bodies are
  abbreviated; ConfigVoc encodes them as expression trees.}
  \label{lst:constraints}
\begin{MintedVerbatim}[commandchars=\\\{\}]
\PYG{c}{\PYGZsh{} Hard constraint (physical limit) \PYGZhy{}\PYGZhy{} existing ConfigVoc}
\PYG{n+nn}{flexi}\PYG{p}{:}\PYG{n+nt}{fuseCapacity} \PYG{k+kt}{a} \PYG{n+nn}{cv}\PYG{p}{:}\PYG{n+nt}{FormulaConstraint} \PYG{p}{;}
  \PYG{n+nn}{cv}\PYG{p}{:}\PYG{n+nt}{usesFeature} \PYG{n+nn}{flexi}\PYG{p}{:}\PYG{n+nt}{sumChargerPower}\PYG{p}{,} \PYG{n+nn}{flexi}\PYG{p}{:}\PYG{n+nt}{fuseCapacity} \PYG{p}{;}
  \PYG{n+nn}{cv}\PYG{p}{:}\PYG{n+nt}{formula} \PYG{p}{[} \PYG{p}{...} \PYG{l+s}{\PYGZdq{}sumChargerPower \PYGZlt{}= fuseCapacity\PYGZdq{}} \PYG{p}{...} \PYG{p}{]} \PYG{p}{.}  \PYG{c}{\PYGZsh{} tree \PYGZhy{} abbreviated here}

\PYG{c}{\PYGZsh{} Soft constraint (user preference) \PYGZhy{}\PYGZhy{} proposed extension}
\PYG{n+nn}{flexi}\PYG{p}{:}\PYG{n+nt}{desiredSOC} \PYG{k+kt}{a} \PYG{n+nn}{cv}\PYG{p}{:}\PYG{n+nt}{FormulaConstraint} \PYG{p}{;}
  \PYG{n+nn}{cv}\PYG{p}{:}\PYG{n+nt}{usesFeature} \PYG{n+nn}{flexi}\PYG{p}{:}\PYG{n+nt}{actualSOC}\PYG{p}{,} \PYG{n+nn}{flexi}\PYG{p}{:}\PYG{n+nt}{targetSOC} \PYG{p}{;}
  \PYG{n+nn}{cv}\PYG{p}{:}\PYG{n+nt}{formula}  \PYG{p}{[} \PYG{p}{...} \PYG{l+s}{\PYGZdq{}actualSOC \PYGZgt{}= targetSOC\PYGZdq{}} \PYG{p}{...} \PYG{p}{]} \PYG{p}{;}          \PYG{c}{\PYGZsh{} tree \PYGZhy{} abbreviated here}
  \PYG{n+nn}{pref}\PYG{p}{:}\PYG{n+nt}{constraintType} \PYG{n+nn}{pref}\PYG{p}{:}\PYG{n+nt}{Soft} \PYG{p}{;}                             \PYG{c}{\PYGZsh{} proposed}
  \PYG{n+nn}{pref}\PYG{p}{:}\PYG{n+nt}{priority} \PYG{l+m+mi}{1000} \PYG{p}{.}                                        \PYG{c}{\PYGZsh{} proposed}

\PYG{n+nn}{flexi}\PYG{p}{:}\PYG{n+nt}{evOwnerPreferences} \PYG{k+kt}{a} \PYG{n+nn}{pref}\PYG{p}{:}\PYG{n+nt}{PreferenceAggregate} \PYG{p}{;}         \PYG{c}{\PYGZsh{} proposed}
  \PYG{n+nn}{pref}\PYG{p}{:}\PYG{n+nt}{aggregates} \PYG{n+nn}{flexi}\PYG{p}{:}\PYG{n+nt}{desiredSOC} \PYG{p}{.}
\end{MintedVerbatim}

\end{listing}

\subsection{Consistency Checking and the Feedback Loop}
\label{sec:loop}

Acquiring a usable constraint set involves relaxation at two distinct points, which
must be distinguished. The first is the \emph{static satisfiability check} of
Figure~\ref{fig:architecture}, run as
constraints are acquired: for each soft preference it asks whether that preference
could be honored \emph{at all} given the hard limits, before any
schedule is computed. Because soft constraints are relaxable, such a preference
does not make the set inconsistent; it simply can never be met.
Operationally the check treats the candidate preference as if
it were hard and tests it against the hard set, which exposes two kinds of clash.
One is \emph{structural}: a preference may contradict the product model or a hard
constraint at the ontological level, such as a request for a forbidden option,
which description-logic reasoning decides. The
other is \emph{numeric}: a preference may violate a value bound outright, as in the
90\,kWh request above, which description logic cannot decide and which requires
reasoning over the arithmetic of the formula constraints. Where this combined check
runs depends on where the preference extension is hosted. On the axiom-based
ConfigDL, an OWL DL reasoner decides the structural case while DL-safe rules in the
Semantic Web Rule Language (SWRL) with arithmetic built-ins check the numeric formula constraints over
already-grounded values (feasibility over open variables is beyond them), keeping
the check inside OWL, though relaxable constraints have no native axiomatic home
there.
On the instance-based ConfigVoc, the formula constraints export to a constraint
solver such as MiniZinc (a path CONTO already provides~\cite{conto-semantics25}), which checks them
numerically and, through reified constraints aggregated into a weighted objective,
gives the soft-constraint semantics a natural target; ASP is a further
transformation target, realizing priorities directly as weak constraints with a
weight and a level. Either way, passing the
static check guarantees only that each preference is individually satisfiable, not
that all can be met together.
Soft preferences are deliberately not tested against one another: because each may
be relaxed, no pair of them is unsatisfiable, so which of two competing preferences
yields is a question of priority for the scheduler, not of consistency.

At this stage, acquisition is a two-way exchange between the neural and symbolic
components, mediated by the ontology. Each assistant grounds an open-ended
natural-language utterance into typed constraints over the shared vocabulary,
so the check operates not on text but on the same formal representation the
hardware sources populate. Because that grounding is fallible, the coupling has a
useful side effect: a mis-grounded constraint (one the assistant translated
incorrectly, say mapping a requested energy to the wrong feature or unit) that
clashes with the hard limits is caught by the same check rather than passing
silently downstream. When a preference
fails the check, it returns a
\emph{justification}: a minimal set of the hard constraints responsible for the
clash.
Both hosting paths supply one, and this is what
selects them: an OWL DL reasoner with SWRL support (Pellet/Openllet) computes the
justification of an inconsistency, the minimal set of clashing axioms and rules, and
a constraint solver such as MiniZinc reports the analogous minimal unsatisfiable
subset of the numeric constraints. A tool that returned only a verdict, without the
clashing set, could not drive the loop. Either way the
result is a set of \emph{named} constraints in the shared vocabulary, not free text. It is handed
back to the assistant that supplied the offending preference as a structured
tool-call result, and the assistant is prompted to surface those named
constraints to the user, renegotiate%
, and feed a revised
constraint into the loop.
The assistant's next turn is thus conditioned on a
symbolic artifact rather than on retrieved text or a scalar signal: the
justification determines which constraint is questioned and how the trade-off is
framed. Iterating this loop restores a statically feasible set.

The second relaxation point is the scheduler, which resolves quantitative and
resource feasibility. Even a structurally consistent set can be jointly
unsatisfiable over the planning horizon, as in the 16:00 request above. Here the
priorities and weights determine which soft constraints are relaxed, and by how
much, so as to minimize total violation cost while the hard constraints hold.
The outcome is reported back to the user through the same assistant,
closing the loop a second time at the level of the realized schedule rather than
the constraint set.

\section{Use Case: The Microgrid}
\label{sec:usecase}

The Siemens microgrid is a use case in the FLEXI project.
Downstream of acquisition, a scheduler
forecasts variable factors such as PV production and grid energy CO\textsubscript{2}
intensity and seeks a schedule that satisfies all hard constraints and as many
soft constraints as possible; making that scheduler neurosymbolic is outside the
scope of this paper.

Constraints are acquired through two assistants. An \emph{EV-owner assistant}
elicits user preferences such as desired departure time and desired energy or
state of charge, and, where legacy hardware cannot report it automatically,
prompts the user for the current state of charge or battery size. An \emph{operator assistant}
captures infrastructure-level requirements from the charge-point operator, such as
stationary-battery reserves and campus power limits. Hardware specifications add
the remaining physical limits: fuse and transformer capacities, battery safety
parameters, and committed flexibility-market obligations (e.g., day-ahead bids).

\paragraph{A worked example.}
The two relaxation points appear in turn. Suppose an EV owner requests an option
the product model forbids for the installed hardware (a structural conflict), or
requests to charge 90\,kWh into a battery that holds 60 (a numeric one). Either way
the preference cannot be honored under the hard limits no matter how anything else
is scheduled; the static check flags it, and the returned justification lets
the assistant ask the user to revise the request before any scheduling is
attempted. Now suppose instead the request is individually feasible, a full charge
by 16:00, but that morning many vehicles charge simultaneously and the stationary
battery is drawn down, so the transformer limit and the battery's reserved power leave too little
capacity to meet every EV owner's target in time. No single request is infeasible,
so the static check passes; the shortfall surfaces only in the scheduler, which
cannot satisfy all soft preferences jointly. Guided by the priorities, it relaxes
the lower-priority ones, delivering an 80\% charge by 16:00, while
holding the hard limits, and reports the outcome to the affected owner.

\section{Discussion and Conclusion}
\label{sec:discussion}

We have argued that multi-stakeholder \emph{constraint acquisition} is a
neurosymbolic problem worth treating in its own right, and instantiated an
ontology-mediated architecture for it on the FLEXI microgrid. Nothing in the
architecture, however, is specific to energy. It applies wherever a
specification must be assembled from constraints acquired across several human
and automated sources of unequal authority, with the ontology as the shared
representation, the hard/soft distinction marking which constraints may be
relaxed, and the explanation-and-renegotiation loop resolving the conflicts that
arise during acquisition. Multi-tenant building climate control is one such
setting, far from energy scheduling yet structurally identical: tenants express
comfort preferences (soft) through their own assistants, a facility manager
contributes building-wide energy caps and safety requirements of higher authority,
and equipment datasheets fix actuator and plant limits (hard); tenants
compete with one another for the same capped supply, as EV owners do for
transformer capacity.

\paragraph{Limitations and open questions.}
We describe an architecture rather than an evaluated system,
and several questions remain open. The preference extension and the explanation
feedback loop are proposed: the extension and the assistants are not yet built, and
the loop has been exercised only on the OWL path. A first evaluation
would target the acquisition loop rather than schedule quality: how often the
static check catches a preference that could never be satisfied, how many
renegotiation turns follow, and whether checking and solver encoding stay within
interactive time. A first design
question is where the preference semantics belong, the ConfigVoc-versus-ConfigDL
choice detailed in Section~\ref{sec:loop}; because both paths supply inconsistency
justifications, it governs how preferences are checked, not whether the loop can
run. Modeling
preferences as weighted soft constraints further raises the questions of how to
set and compare priorities across stakeholder groups, and how to handle multiple
competing soft constraints (multi-objective optimization); because the groups are
self-interested, priority elicitation also carries a social-choice dimension, with
the risk of strategic misreporting. The approach also assumes the LLM assistants
translate faithfully between natural language and the ontology vocabulary; a
mis-grounding that clashes with the hard limits is caught, but a consistent yet
wrong one would propagate silently, making faithful grounding the make-or-break
assumption.
For grounding we would adopt existing practice rather than extend it: constrained
decoding against the vocabulary, as in the configuration
copilot~\cite{conto-copilot}, which yields well-formed constraints but not
necessarily the intended ones. Furthermore, description-logic reasoning must remain
fast enough for interactive renegotiation. Resolving the ConfigVoc-versus-ConfigDL
placement and establishing priority semantics, faithful elicitation, and
interactive-time reasoning is the research agenda this work motivates; a first
prototype of the preference extension is our immediate next step.

\begin{acknowledgments}
This research project is funded by CETPartnership, the Clean Energy Transition Partnership under the 2024 joint call for research proposals, co-funded by the European Commission (GA N°101069750) and with the funding organisations FFG Austrian Research Promotion Agency (Austria), NWO (Dutch Research Council) (the Netherlands), Swedish Energy Agency (Sweden) and GSRI (Greece).
\end{acknowledgments}

\section*{Declaration on Generative AI}
During the preparation of this work, the authors used Claude in order to: Paraphrase and reword, Improve writing style, Grammar and spelling check. 
After using this tool, the authors reviewed and edited the content as needed and take full responsibility for the publication's content.

\bibliography{sample-ceur}

\begin{thebibliography}{14}
\expandafter\ifx\csname natexlab\endcsname\relax\def\natexlab#1{#1}\fi
\providecommand{\url}[1]{\texttt{#1}}
\providecommand{\href}[2]{#2}
\providecommand{\path}[1]{#1}
\providecommand{\DOIprefix}{doi:}
\providecommand{\ArXivprefix}{arXiv:}
\providecommand{\URLprefix}{URL: }
\providecommand{\Pubmedprefix}{pmid:}
\providecommand{\doi}[1]{\href{http://dx.doi.org/#1}{\path{#1}}}
\providecommand{\Pubmed}[1]{\href{pmid:#1}{\path{#1}}}
\providecommand{\bibinfo}[2]{#2}
\ifx\xfnm\relax \def\xfnm[#1]{\unskip,\space#1}\fi
\bibitem[{van Harmelen and ten Teije(2019)}]{boxology}
\bibinfo{author}{F.~van Harmelen}, \bibinfo{author}{A.~ten Teije},
\newblock \bibinfo{title}{A boxology of design patterns for hybrid learning and
  reasoning systems},
\newblock \bibinfo{journal}{Journal of Web Engineering} \bibinfo{volume}{18}
  (\bibinfo{year}{2019}) \bibinfo{pages}{97--124}. \URLprefix
  \url{https://ieeexplore.ieee.org/document/10247288}.
\bibitem[{Scholak et~al.(2021)Scholak, Schucher, and Bahdanau}]{picard}
\bibinfo{author}{T.~Scholak}, \bibinfo{author}{N.~Schucher},
  \bibinfo{author}{D.~Bahdanau},
\newblock \bibinfo{title}{{PICARD}: Parsing incrementally for constrained
  auto-regressive decoding from language models},
\newblock in: \bibinfo{booktitle}{Proceedings of the Conference on Empirical
  Methods in Natural Language Processing ({EMNLP})}, \bibinfo{year}{2021}, pp.
  \bibinfo{pages}{9895--9901}. \DOIprefix\doi{10.18653/v1/2021.emnlp-main.779}.
\bibitem[{Pan et~al.(2023)Pan, Albalak, Wang, and Wang}]{logiclm}
\bibinfo{author}{L.~Pan}, \bibinfo{author}{A.~Albalak},
  \bibinfo{author}{X.~Wang}, \bibinfo{author}{W.~Y. Wang},
\newblock \bibinfo{title}{{Logic-LM}: Empowering large language models with
  symbolic solvers for faithful logical reasoning},
\newblock in: \bibinfo{booktitle}{Findings of the Association for Computational
  Linguistics: {EMNLP}}, \bibinfo{year}{2023}, pp. \bibinfo{pages}{3806--3824}.
  \DOIprefix\doi{10.18653/v1/2023.findings-emnlp.248}.
\bibitem[{Olausson et~al.(2023)Olausson, Gu, Lipkin, Zhang, Solar-Lezama,
  Tenenbaum, and Levy}]{linc}
\bibinfo{author}{T.~X. Olausson}, \bibinfo{author}{A.~Gu},
  \bibinfo{author}{B.~Lipkin}, \bibinfo{author}{C.~E. Zhang},
  \bibinfo{author}{A.~Solar-Lezama}, \bibinfo{author}{J.~B. Tenenbaum},
  \bibinfo{author}{R.~Levy},
\newblock \bibinfo{title}{{LINC}: A neurosymbolic approach for logical
  reasoning by combining language models with first-order logic provers},
\newblock in: \bibinfo{booktitle}{Proceedings of the Conference on Empirical
  Methods in Natural Language Processing ({EMNLP})}, \bibinfo{year}{2023}, pp.
  \bibinfo{pages}{5153--5176}. \DOIprefix\doi{10.18653/v1/2023.emnlp-main.313}.
\bibitem[{Ye et~al.(2023)Ye, Chen, Dillig, and Durrett}]{satlm}
\bibinfo{author}{X.~Ye}, \bibinfo{author}{Q.~Chen},
  \bibinfo{author}{I.~Dillig}, \bibinfo{author}{G.~Durrett},
\newblock \bibinfo{title}{{SatLM}: Satisfiability-aided language models using
  declarative prompting},
\newblock in: \bibinfo{booktitle}{Advances in Neural Information Processing
  Systems ({NeurIPS})}, volume~\bibinfo{volume}{36}, \bibinfo{year}{2023}, pp.
  \bibinfo{pages}{45548--45580}. \URLprefix
  \url{https://proceedings.neurips.cc/paper_files/paper/2023/hash/8e9c7d4a48bdac81a58f983a64aaf42b-Abstract-Conference.html}.
\bibitem[{Bischof et~al.(2025)Bischof, Falkner, Filtz, Schneider, Steyskal, and
  Topa}]{conto-semantics25}
\bibinfo{author}{S.~Bischof}, \bibinfo{author}{A.~Falkner},
  \bibinfo{author}{E.~Filtz}, \bibinfo{author}{P.~Schneider},
  \bibinfo{author}{S.~Steyskal}, \bibinfo{author}{M.-L. Topa},
\newblock \bibinfo{title}{{CONTO}: An ontology-based approach for interoperable
  configuration knowledge},
\newblock in: \bibinfo{editor}{D.~Chaves-Fraga}, \bibinfo{editor}{I.~Heibi},
  \bibinfo{editor}{D.~Garijo}, \bibinfo{editor}{D.~Collarana},
  \bibinfo{editor}{A.~Salatino}, \bibinfo{editor}{S.~Vahdati} (Eds.),
  \bibinfo{booktitle}{Posters and Demos Track of {SEMANTiCS}}, volume
  \bibinfo{volume}{4064} of \textit{\bibinfo{series}{CEUR Workshop
  Proceedings}}, \bibinfo{year}{2025}. \URLprefix
  \url{https://ceur-ws.org/Vol-4064/PD-paper13.pdf}.
\bibitem[{Bessiere et~al.(2017)Bessiere, Koriche, Lazaar, and
  O'Sullivan}]{constraint-acquisition}
\bibinfo{author}{C.~Bessiere}, \bibinfo{author}{F.~Koriche},
  \bibinfo{author}{N.~Lazaar}, \bibinfo{author}{B.~O'Sullivan},
\newblock \bibinfo{title}{Constraint acquisition},
\newblock \bibinfo{journal}{Artificial Intelligence} \bibinfo{volume}{244}
  (\bibinfo{year}{2017}) \bibinfo{pages}{315--342}.
  \DOIprefix\doi{10.1016/j.artint.2015.08.001}.
\bibitem[{Bessiere et~al.(2013)Bessiere, Coletta, Hebrard, Katsirelos, Lazaar,
  Narodytska, Quimper, and Walsh}]{quacq}
\bibinfo{author}{C.~Bessiere}, \bibinfo{author}{R.~Coletta},
  \bibinfo{author}{E.~Hebrard}, \bibinfo{author}{G.~Katsirelos},
  \bibinfo{author}{N.~Lazaar}, \bibinfo{author}{N.~Narodytska},
  \bibinfo{author}{C.-G. Quimper}, \bibinfo{author}{T.~Walsh},
\newblock \bibinfo{title}{Constraint acquisition via partial queries},
\newblock in: \bibinfo{booktitle}{Proceedings of the International Joint
  Conference on Artificial Intelligence ({IJCAI})}, \bibinfo{year}{2013}, pp.
  \bibinfo{pages}{475--481}. \URLprefix
  \url{https://www.ijcai.org/Proceedings/13/Papers/078.pdf}.
\bibitem[{Zhang et~al.(2024)Zhang, Carriero, Schreiberhuber, Tsaneva,
  S\'anchez~Gonz\'alez, Kim, and de~Berardinis}]{ontochat}
\bibinfo{author}{B.~Zhang}, \bibinfo{author}{V.~A. Carriero},
  \bibinfo{author}{K.~Schreiberhuber}, \bibinfo{author}{S.~Tsaneva},
  \bibinfo{author}{L.~S\'anchez~Gonz\'alez}, \bibinfo{author}{J.~Kim},
  \bibinfo{author}{J.~de~Berardinis},
\newblock \bibinfo{title}{{OntoChat}: A framework for conversational ontology
  engineering using language models},
\newblock in: \bibinfo{booktitle}{The Semantic Web: {ESWC} Satellite Events},
  \bibinfo{year}{2024}, pp. \bibinfo{pages}{102--121}.
  \DOIprefix\doi{10.1007/978-3-031-78952-6_10}.
\bibitem[{Lawless et~al.(2024)Lawless, Schoeffer, Le, Rowan, Sen, St.~Hill,
  Suh, and Sarrafzadeh}]{lawless-elicitation}
\bibinfo{author}{C.~Lawless}, \bibinfo{author}{J.~Schoeffer},
  \bibinfo{author}{L.~Le}, \bibinfo{author}{K.~Rowan},
  \bibinfo{author}{S.~Sen}, \bibinfo{author}{C.~St.~Hill},
  \bibinfo{author}{J.~Suh}, \bibinfo{author}{B.~Sarrafzadeh},
\newblock \bibinfo{title}{``{I} want it that way'': Enabling interactive
  decision support using large language models and constraint programming},
\newblock \bibinfo{journal}{{ACM} Transactions on Interactive Intelligent
  Systems} \bibinfo{volume}{14} (\bibinfo{year}{2024})
  \bibinfo{pages}{22:1--22:33}. \DOIprefix\doi{10.1145/3685053}.
\bibitem[{Kogler et~al.(2024)Kogler, Chen, Falkner, Haselb\"ock, and
  Wallner}]{conto-copilot}
\bibinfo{author}{P.~Kogler}, \bibinfo{author}{W.~Chen}, \bibinfo{author}{A.~A.
  Falkner}, \bibinfo{author}{A.~Haselb\"ock}, \bibinfo{author}{S.~Wallner},
\newblock \bibinfo{title}{Configuration copilot: Towards integrating large
  language models and constraints},
\newblock in: \bibinfo{editor}{E.~Vareilles}, \bibinfo{editor}{C.~Grosso},
  \bibinfo{editor}{J.~M. Horcas}, \bibinfo{editor}{A.~Felfernig} (Eds.),
  \bibinfo{booktitle}{Proceedings of the International Workshop on
  Configuration ({ConfWS})}, volume \bibinfo{volume}{3812} of
  \textit{\bibinfo{series}{CEUR Workshop Proceedings}}, \bibinfo{year}{2024},
  pp. \bibinfo{pages}{101--110}. \URLprefix
  \url{https://ceur-ws.org/Vol-3812/paper14.pdf}.
\bibitem[{Schiex et~al.(1995)Schiex, Fargier, and Verfaillie}]{vcsp}
\bibinfo{author}{T.~Schiex}, \bibinfo{author}{H.~Fargier},
  \bibinfo{author}{G.~Verfaillie},
\newblock \bibinfo{title}{Valued constraint satisfaction problems: Hard and
  easy problems},
\newblock in: \bibinfo{booktitle}{Proceedings of the International Joint
  Conference on Artificial Intelligence ({IJCAI})}, \bibinfo{year}{1995}, pp.
  \bibinfo{pages}{631--639}. \URLprefix
  \url{https://www.ijcai.org/Proceedings/95-1/Papers/083.pdf}.
\bibitem[{Bistarelli et~al.(1997)Bistarelli, Montanari, and
  Rossi}]{semiring-csp}
\bibinfo{author}{S.~Bistarelli}, \bibinfo{author}{U.~Montanari},
  \bibinfo{author}{F.~Rossi},
\newblock \bibinfo{title}{Semiring-based constraint satisfaction and
  optimization},
\newblock \bibinfo{journal}{J. ACM} \bibinfo{volume}{44} (\bibinfo{year}{1997})
  \bibinfo{pages}{201--236}. \DOIprefix\doi{10.1145/256303.256306}.
\bibitem[{Buccafurri et~al.(2000)Buccafurri, Leone, and
  Rullo}]{weak-constraints}
\bibinfo{author}{F.~Buccafurri}, \bibinfo{author}{N.~Leone},
  \bibinfo{author}{P.~Rullo},
\newblock \bibinfo{title}{Enhancing disjunctive datalog by constraints},
\newblock \bibinfo{journal}{{IEEE} Trans. Knowl. Data Eng.}
  \bibinfo{volume}{12} (\bibinfo{year}{2000}) \bibinfo{pages}{845--860}.
  \DOIprefix\doi{10.1109/69.877512}.

\end{thebibliography}

\end{document}